\documentclass{article}

\usepackage[preprint, nonatbib]{neurips_2021}

\usepackage{amsmath} 
\usepackage{amssymb} 

\usepackage{subfigure}

\usepackage[utf8]{inputenc} 
\usepackage[T1]{fontenc}    
\usepackage{hyperref}       
\usepackage{url}            
\usepackage{booktabs}       
\usepackage{amsfonts}       
\usepackage{nicefrac}       
\usepackage{microtype}      
\usepackage{xcolor}         
\usepackage{graphicx}

\usepackage{cite}

\title{Deep Reinforcement Learning Models Predict Visual Responses in the Brain: A Preliminary Result}

\author{
  \textbf{Maytus Piriyajitakonkij $^{1, 3}$\thanks{Work done during working at VISTEC.} , Sirawaj Itthipuripat $^2$} \\ 
  \textbf{Theerawit Wilaiprasitporn $^1$, Nat Dilokthanakul $^1$}
  \\
  $^1$ Vidyasirimedhi Institute of Science and Technology (VISTEC), Thailand \\ 
  $^2$ King Mongkut's University of Technology Thonburi (KMUTT), Thailand \\
  $^3$ Imperial College London, United Kingdom \\
  \texttt{natd\_pro@vistec.ac.th} 
}

\begin{document}

\maketitle

\begin{abstract}
Supervised deep convolutional neural networks (DCNNs) are currently one of the best computational models that can explain how the primate ventral visual stream solves object recognition. However, embodied cognition has not been considered in the existing visual processing models. From the ecological standpoint, humans learn to recognize objects by interacting with them, allowing better classification, specialization, and generalization. Here, we ask if computational models under the embodied learning framework can explain mechanisms underlying object recognition in the primate visual system better than the existing supervised models? To address this question, we use reinforcement learning to train neural network models to play a 3D computer game and we find that these reinforcement learning models achieve neural response prediction accuracy scores in the early visual areas (e.g., V1 and V2) in the levels that are comparable to those accomplished by the supervised neural network model. In contrast, the supervised neural network models yield better neural response predictions in the higher visual areas, compared to the reinforcement learning models. Our preliminary results suggest the future direction of visual neuroscience in which deep reinforcement learning should be included to fill the missing embodiment concept.
\end{abstract}

\section{Introduction}
\label{introduction}

The primate visual system has a remarkable ability in recognizing multiple classes of visual objects and scenes. Object identification and recognition are thought to be supported in part by the ventral visual stream, which comprises multiple cortical areas resided in the occipital and the temporal lobes. These visual areas are organized in a hierarchical fashion, where higher visual areas process more complex visual features. The primary visual cortex (V1) is thought to encode
low-level features, such as stimulus contrasts, orientations, edges and spatial frequencies. The visual inputs from V1 are then relayed through the extrastriate visual areas like V2 and V4, thought to process mid-level visual features such as textures, colors, and motions \cite{movshon1978spatial, roe2012toward, freeman2013functional}, and ultimately projected onto the downstream visual area known as the inferior temporal (IT) cortex, where neural representations of different object categories can be untangled or linearly separated \cite{dicarlo2012does}.
 

Over decades, several computational models have been proposed to explain how the ventral visual stream solves object recognition. Among these models, deep convolutional neural network (DCNN) is currently one of the most promising models that can explain neural mechanisms underlying object-related information processing in the ventral visual stream. Specifically, when the supervised DCNN is trained with natural images to minimize the classification error, its activation can be used to predict patterns of behavioral and neural responses in the primate ventral visual stream that are related to particular sets of objects better than other existing models \cite{yamins2014performance, yamins2016using, rajalingham2018large}, e.g., V1-like Gabor-based model \cite{pinto2008real} and HMAX \cite{serre2007feedforward}.


However, embodied visual cognition has not been considered in the previous studies. The brain does not only perceive and process information, it also interacts with the physical world and changes the state of what it perceives \cite{gibson1977theory}. Embodied cognition helps the brain make sense of the three-dimensional world from two-dimensional sensory information. For example, it helps the brain infer the depth of an object \cite{gibson1979ecological,gibson1988exploratory, wilson2002six}. We argue that computational models of a biological visual system must be embodied. 

\begin{figure}
\vskip 0in
\begin{center}
\centerline{\includegraphics[width=1\columnwidth]{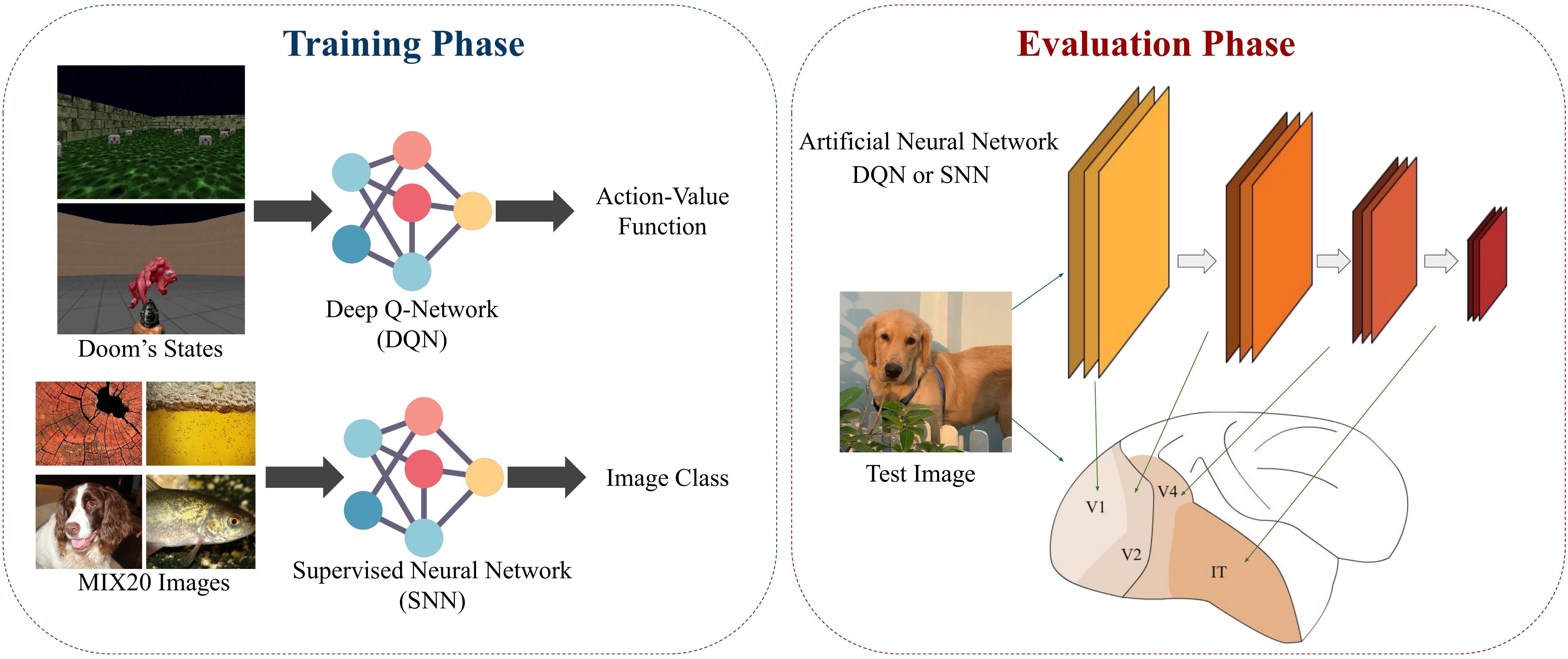}}
\caption{\textbf{Training Phase:} DQN models are trained to accomplish tasks in Doom environment. SNN models are trained to minimize classification errors. \textbf{Evaluation Phase:} The neural network model and the brain receive the same images (blue arrows). Each model-layer's activation is mapped by linear transformation to neural responses in each brain area (green arrows). Model activation is used to predict neural responses in the area to which its layer is mapped.
}
\label{brain_model}
\end{center}
\vskip -0.4in
\end{figure}

Reinforcement learning (RL) is one way to create an embodied agent by using reward to shape action selections. The agent learns to select sequences of actions that maximize the total reward \cite{sutton2018reinforcement}. Deep RL leverages a representational advantage of deep neural networks and significantly improve RL ability to solve much more complex problems such as playing Atari games with raw pixels, winning the Go's world champion and robotic control \cite{mnih2013playing, lillicrap2015continuous, silver2016mastering}.



In this work, we use deep Q-learning--a deep RL algorithm--to train DCNN \cite{mnih2013playing}. The neural network model is trained to play a 3D computer game. It receives only raw pixels to take action and accomplish tasks in the game. We hypothesize that visual features learned by DQN are similar to those learned by the ventral visual stream. We explore this hypothesis in our experiments, and find that DQN achieves comparable neural prediction accuracy scores to supervised learning model in V1 and V2 using the same neural network architecture.  


\section{Method}
The overviews of model training and evaluation are shown in \autoref{brain_model}. At the training phase, artificial neural network (ANN) models are trained with supervised and reinforcement learning paradigms. DQN models are trained to play Doom--a 3D shooting game. Supervised Neural Network (SNN) models are trained to recognize images. 

At the evaluation phase, these models are evaluated with neural data by their neural predictivity scores. Neural predictivity is used to measure how well activations in the neural network model predict neural responses of visual stimuli. The set of images and neural responses, used for evaluation, are explained in \autoref{neural_data}. The detail of neural predictivity, DQN, RL environment/datasets and model configuration are provided in \autoref{neural_pred}, \autoref{dqn}, \autoref{environment} and \autoref{model_config} respectively.

\subsection{Neural Data}
\label{neural_data}
There are two evaluation datasets: V1/V2 dataset and V4/IT dataset. Neural responses of V1 and V2 areas were collected by \cite{freeman2013functional} from 13 macaques with implanted electrodes. Neural data were recorded from 102 and 103 neurons in V1 and V2 respectively. Stimuli were from 15 texture families. Neural responses of V4 and IT areas were collected by \cite{majaj2015simple} from two macaques with implanted electrodes, recorded from 88 neurons in V4 and 168 neurons in IT. There were a total of 2,560 stimuli from 64 rendered three-dimensional objects from 8 categories with the random pose, position, size, and natural background.
\subsection{Neural Predictivity}
\label{neural_pred}
At the evaluation phase, the brain and the model receives input images from evaluation datasets. The pairs of neural responses $y$ and model activations $x$ are attained corresponding to the same input images.
Neural predictivity indicates a prediction performance of a linear regression model, which is trained to predict neural responses $y$ from model activations $x$. In case of DCNN, activations of each layer are used to predict each brain area neural responses as shown in \autoref{brain_model}. Predicted response $y'$ is derived from linear transformation $y' = wx$, where w is a parameter vector of the linear model. Then, neural predictivity score is computed using \autoref{eq:correlation}.

\begin{equation}\label{eq:correlation}
r = \frac{\sum_{i=1}^{n}(y_{i}-\bar{y})(y_{i}'-\bar{y}')}{\sqrt{\sum_{i=1}^{n}(y_{i}-\bar{y})^{2}(y_{i}'-\bar{y}')^{2}}}
\end{equation}

Where $\bar{y}$ and $\bar{y}'$ are averages of $y$ and $y'$ across samples respectively. Here, we use \emph{Brain-Score} library \cite{schrimpf2018brain, schrimpf2020integrative} to implement this process.

\subsection{Deep Q-Networks}
\label{dqn}
Q-learning is a learning method that learns the value of a state-action pair, i.e. how good is an action $a$ if taken in a state $s$. Deep Q-Network (DQN) is an extension to the Q-learning algorithm.
It leverages the representational power of a convolutional neural network to represent the state-action value or Q-value. In other word, it uses a deep neural network $\theta$ as a function approximator $Q(s,a; \theta)$, which maps a state-action pair into a scalar value. Therefore, the early layers in the network $\theta$ have to be able to extract useful features for estimating the Q-value. 
DQN can be trained with the following objective,
\begin{equation}\label{eq:loss}
L_{i}(\theta_{i}) = \mathbb{E}_{s,a,s'}[(y_{i}-Q(s,a;\theta_{i}))^{2}]
\end{equation}
where the target $y_{i} = r+ \gamma \text{max}_{a'}Q(s',a';\theta_{i-1})$, $r$ is reward given by an environment, and $s'$ and $a'$ are a state and an action in the next time-step observed during the interactions with the environment.
In the experiments, we follows the dueling architecture by Wang et al. \cite{wang2016dueling} 

\subsection{Environment and Dataset}
\label{environment}
We have eight trained models. Six models are DQN models trained on different RL tasks. The other two models are supervised models (SNN). All of the models have the same architecture except for the output size. One of the SNNs is trained with CIFAR10 \cite{krizhevsky2009learning}, containing 10 classes of natural images. Another is trained with our own MIX20 dataset, which has more image variations and more texture information than CIFAR10.
The details of RL tasks and MIX20 are explained below.

\textbf{Environment:} DQN is an embodied agent. It needs the environment to interact with during training. However, during predictivity evaluation, we can get the model activations by directly pass any stimuli images unrelated to the environment through the model.

We use \emph{VizDoom} platform \cite{kempka2016vizdoom} (\url{https://github.com/mwydmuch/ViZDoom}) for our training environments. Six \emph{VizDoom} tasks are chosen. \emph{Simpler Basic}, \emph{Defend Center} and \emph{Predict Position} are monster shooting tasks. \emph{Predict Position} and \emph{Take Cover} are missile motion learning tasks. \emph{My Way Home} is a visual navigation task in labyrinth layout. \emph{Health Gathering} is an object collection task. The set of actions an agent can take are shoot, turn left, turn right, move left, move right, move forward. The agent is allowed to take some of these actions depending on the tasks.

\textbf{MIX20 dataset:} The visual stimuli in V1/V2 dataset are texture images, and in V4/IT dataset are natural object images. To create a best possible model, we want similar image statistics in our dataset and in the evaluation datasets. Therefore, we construct 20-class image dataset (MIX20) from 10 objects in ImageNet dataset \cite{russakovsky2015imagenet} and from 10 textures from DTD dataset \cite{cimpoi2014describing}. The baseline trained with this dataset serves as our upper-bound, a cheat model that has been trained on a relatively better dataset. 


\subsection{Model}
\label{model_config}
The models' inputs are 128$\times$128-pixel color images. The models that are used for all learning paradigms have almost the same architecture. The only exception in the number of output nodes which depends on an action space or image classes. The architecture has four consecutive convolutional layers, one fully-connected, and a output layer. The number of features map are 16, 32, 64 and 32 respectively. Kernel size is 3 for all layers. Stride lengths are 2, 2, 1 and 1 respectively. The fully-connected layer has 64 hidden units. ReLu activation function is applied in all layers, and BatchNorm2D \cite{ioffe2015batch} is applied to all convolutional layers. The total number of parameters is about 1.5 million, depending on the number of output nodes.

\section{Experiment}
We compare neural predictivity scores among all models.  
An untrained model, which is randomly initialized, is used as a baseline. 
The untrained model gives us a lower-bound of neural predictivity score. To support our hypothesis, we expect the DQN models to have higher neural predictivity scores than the untrained model.

We also compare the RL-trained models with two supervised models.
The CIFAR10 model represents a supervised model, which is trained on 
a dataset with relatively similar complexity to the DOOM environments. The MIX20 model represents a supervised model, which is trained on a good dataset with 
high texture variations and natural looking images. This model serves as our upper-bound to the neural predictivity score achievable by the small architecture used in this study.



\begin{figure}
\vskip 0in
\begin{center}
\centerline{\includegraphics[width=1.0\columnwidth]{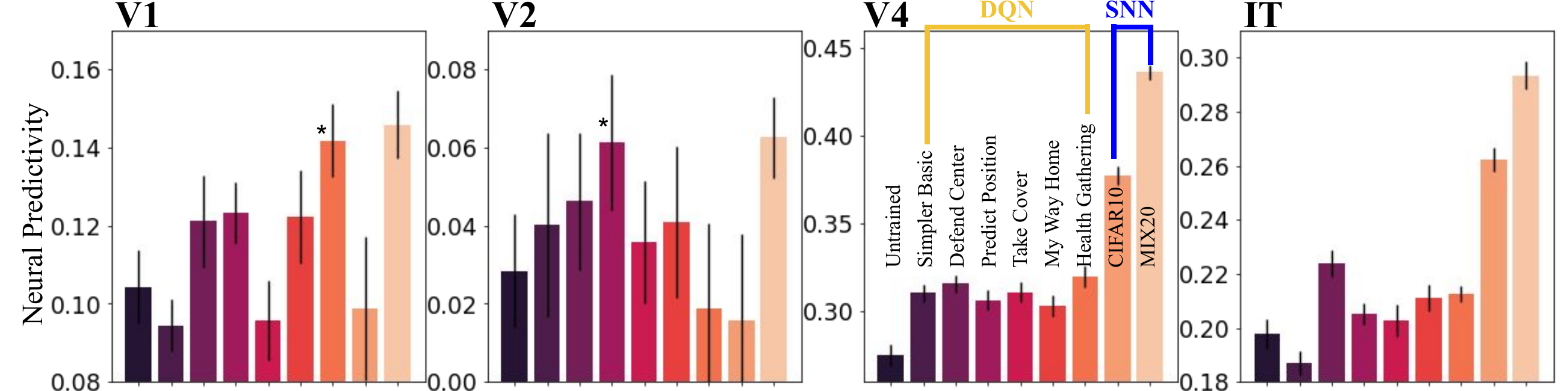}}
\caption{Neural Predictivity scores of the untrained model, DQN and SNN models in four brain areas: V1, V2, V4 and IT. The error bars are computed using standard error.}
\label{result}
\end{center}
\vskip -0.4in
\end{figure}

\section{Result and Discussion}

\autoref{result} shows the average neural predictivity scores of the models in four brain areas. In summary, two DQN models achieve comparable neural predictivity scores to MIX20-trained SNN in V1 and V2 areas. MIX20-trained SNN is the best model for neural prediction in all areas. Moreover, all DQN models achieve higher V4 neural predictivity scores than the untrained model. While CIFAR10-trained SNN is the second-best model for V4 and IT areas, its neural predictivity scores in V1 and V2 are lower than the untrained model.





We see that RL-trained models have high predictivity scores compares to the untrained models. This result supports the hypothesis that reinforcement learning (RL) can shape visual representation in the biological brain and it is especially good at learning low-level features as seen in high predictivity scores in V1 and V2. 

DQN models are only trained in non-photorealistic and simple RL environments. In contrast, SNN is trained with MIX20 which has much richer and more natural visual information. Interestingly, DQN models achieve comparable performance to MIX20-trained SNN in V1 and V2 neural predictions. It is possible that the learning dynamics of RL leads to a closer texture representation to the texture representation in the brain. This needs further investigation and we plan to investigate an RL agent in a more complex and photorealistic environments in the future work.


The DQN models are trained in one layout. Therefore, these models receive images that have similar statistics. As a result, we expect that these models to also learn similar visual features. Nevertheless, the experimental results show that the DQN models, trained from different tasks, have different neural predictivity scores. This implies that they learn different visual features. \emph{Defend-Center}-trained DQN achieves much higher neural predictivity than \emph{Simpler-Basic}-trained DQN neural predictivity in V1 and IT areas. The main difference between \emph{Simpler Basic} and \emph{Defend Center} is that the latter has higher task complexity. We think that the DQN trained to achieve a simple task learn poorer visual features than the more complex tasks.

CIFAR10-trained SNN does not improve neural prediction in V1 and V2 areas from the untrained model. CIFAR10 is a much lower quality dataset than MIX20 dataset. Moreover, our DCNN models have a small number of parameters. Poor images in CIFAR10 dataset and the low model complexity might cause it to learn poor low-level visual features, resulting in a lower neural prediction accuracy in V1 and V2. Interestingly, CIFAR10-trained model performs relatively well in predicting neural responses in V4 and IT. 

All SNN models perform better than all the DQN models in V4 and IT neural predictions. The object identity representations are disentangled in the late SNN layer and in IT cortex. Since SNN models are trained specifically to disentangle object class, it is understandable that they could perform better in IT neural predictivity. DQN models are not directly trained to classify an object. They are trained to achieve higher level goals. It is possible that to achieve the goals in these environments the networks do not require to object disentanglement. In real-world, however, the tasks are much more complex. We also plan to investigate further in this direction.

In summary, we found that RL-trained models has relatively good neural predictivity scores compare to the supervised models, especially in V1 and V2, even though the environments are not photo-realistic. For future work, we plan to use more realistic environments and investigate various kinds of RL algorithms and tasks. Another interesting question is how ventral stream interacts with dorsal stream, which is thought to encode object spatial representations and guidance of actions. Since deep RL models, trained in complex environment, should be able to encode object identity, object spatial locations, and guidance of actions, we think that the encoded information in deep RL models might help explain the interaction between the two streams.

\section*{Acknowledgement}
This work was supported by Thailand Science Research and Innovation (SRI62W1501). The authors would like to thank Chaipat Chunharas and Supanida Piyayotai for their help in generating the preliminary ideas which lead to this work. The authors also would like to thank Wanitchaya Poonpatanapricha for useful discussion on the research problems and ideas in this work.


\end{document}